\documentclass[authoryear,10pt]{article}
\usepackage[margin=2cm]{geometry}
\usepackage{cite}
\usepackage{amsmath,amssymb,amsfonts}
\usepackage{algorithmic}
\usepackage{graphicx}
\usepackage{textcomp}
\usepackage{float}
\usepackage{color}
\usepackage{xcolor}
\usepackage{hyperref}
\usepackage[T1]{fontenc}
\usepackage{tgheros}
\usepackage{parskip}
\usepackage{caption}
\usepackage{subcaption}
\usepackage[capitalise]{cleveref}
\usepackage{booktabs}
\usepackage{multirow}
\usepackage{mathtools}
\usepackage{setspace}
\usepackage[T1]{fontenc}

\hypersetup{
    colorlinks,
    linkcolor={blue!50!black},
    citecolor={blue!50!black},
    urlcolor={blue!50!black}
}

\title{Automatic dataset shift identification to support\\safe deployment of medical imaging AI}
\author{
Mélanie Roschewitz, Raghav Mehta, Charles Jones, Ben Glocker\\
\small{Imperial College London}
}
\date{}

\begin{document}

\maketitle

\begin{abstract}
    Shifts in data distribution can substantially harm the performance of clinical AI models and lead to misdiagnosis. Hence, various methods have been developed to detect the presence of such shifts at deployment time. However, the root causes of dataset shifts are diverse, and the choice of shift mitigation strategies is highly dependent on the precise type of shift encountered at test time. As such, \emph{detecting} test-time dataset shift is not sufficient: precisely \emph{identifying} which type of shift has occurred is critical. In this work, we propose the first unsupervised dataset shift identification framework for imaging datasets, effectively distinguishing between prevalence shift (caused by a change in the label distribution), covariate shift (caused by a change in input characteristics) and mixed shifts (simultaneous prevalence and covariate shifts). We discuss the importance of self-supervised encoders for detecting subtle covariate shifts and propose a novel shift detector leveraging both self-supervised encoders and task model outputs for improved shift detection. We show the effectiveness of the proposed shift identification framework across three different imaging modalities (chest radiography, digital mammography, and retinal fundus images) on five types of real-world dataset shifts using five large publicly available datasets. Code is publicly available at \url{https://github.com/biomedia-mira/shift_identification}.
\end{abstract}

\section{Introduction}
\begin{figure}
    \centering
    \includegraphics[width=\textwidth]{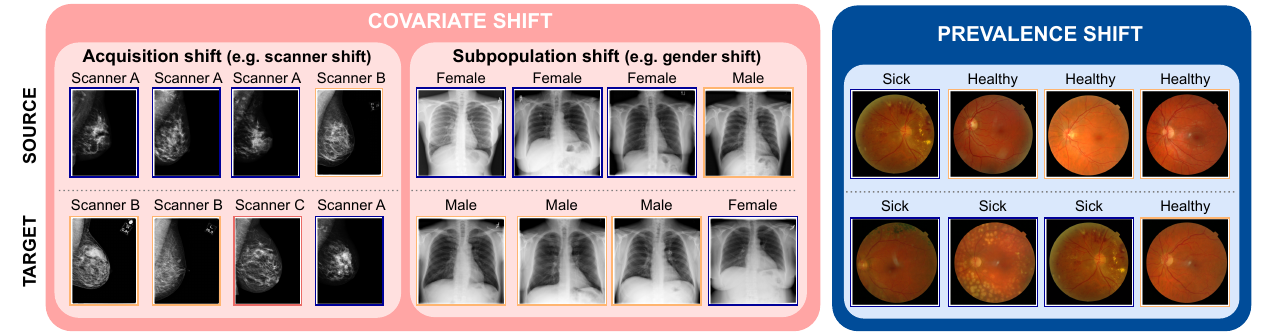}
    \caption[Examples of dataset shifts in medical imaging]{\textbf{Examples of dataset shifts in medical imaging}. Reliably detecting and identifying the nature of the shift is crucial to enable the safe deployment of machine learning systems applications. In this work, we propose the first shift \textit{identification} framework able to reliably detect and classify any detected shift as (i) prevalence, (ii) covariate or (iii) mixed prevalence and covariate shift, for imaging datasets.}
    \label{fig:types_shifts}
\end{figure}

\index{Prevalence shift}
\index{Acquisition shift}
\index{Covariate shift}
\index{Label shift}
Machine learning models are notoriously sensitive to changes in the input data distribution, a phenomenon commonly referred to as dataset shift~\cite{zhou_domain_2023}. This is particularly problematic in clinical settings, where dataset shift is a common occurrence and may arise from various factors~\cite{castro_causality_2020}. Changes in the frequency of disease positives over time or across geographical regions cause \emph{prevalence shift}~\cite{godau_deployment_2023,ma_test-time_2022}. The use of different acquisition protocols or scanners~\cite{yan_mri_2020,sharma_multi-vendor_2023,stacke_measuring_2021}, or a change in patient demographics~\cite{appelman_sex_2015,yang_limits_2024} can induce shifts in image characteristics, known as \emph{covariate shift}. We illustrate examples of real-world shifts in~\cref{fig:types_shifts}. Dataset shift can dramatically affect AI performance of AI and may lead to clinical errors such as misdiagnosis~\cite{seyyed-kalantari_underdiagnosis_2021,sahiner_data_2023,finlayson_clinician_2021}. It is hence crucial to implement safeguards allowing not only effective detection of the \emph{presence} of shifts, but importantly, reliable \emph{identification} of the root causes.

Dataset shifts can be detected at deployment time by using statistical testing to compare the distributions of incoming test data to the distribution of the reference data (representative of the data used to validate the deployed AI model). Significant progress has been made in this field where state-of-the-art methods can detect various types of real-world shifts~\cite{koch_deep_2023,koch_distribution_2024,rabanser_failing_2019,feng_designing_2024}. Shifts between test and reference data can either be detected at the output level (by comparing distributions of model outputs), or at the input level (by comparing low-dimensional feature representations of input images)~\cite{rabanser_failing_2019}. In this work, we show that different types of shifts require different shift detection approaches. On the one hand, comparing model output distributions allows for the reliable detection of shifts directly related to the downstream task, such as changes in prevalence. On the other hand, we show that, for shifts orthogonal to the downstream task, such as changes in image acquisition protocols, comparing output distributions is not sufficient. For such shifts, test and reference data need to be compared at the input level using rich feature representations. We demonstrate that self-supervised neural network image encoders, trained without using any task-specific annotations, yield excellent low-dimensional feature representations for shift detection.

While \emph{detecting} dataset shifts is important, it is insufficient for the safe deployment of AI. Besides knowing that there is a problem, we need to be able to \emph{identify} the precise type of shift to take the necessary actions, implement preventive measures, and safeguard against harm caused by AI errors. Indeed, many domain adaptation techniques are shift-specific: applying the wrong mitigation technique may, in the best case, be ineffective in resolving the shift or, in the worst case, severely harm model performance or calibration. For example, prevalence shifts can often be mitigated with lightweight output recalibration techniques~\cite{alexandari_maximum_2020,wen_class_2024}, but these rely on the assumption that no other types of shift are present. Applying such label shift adaptation methods when the shift is actually caused by covariate shift may drastically degrade model calibration and clinical metrics. In contrast, covariate shifts require more advanced domain adaptation techniques or model finetuning~\cite{zuo_unsupervised_2021,kang_stainnet_2021,xie_learning_2018}. For example, image-harmonisation techniques (e.g. \cite{kang_stainnet_2021}) or automatic correction methods (e.g.~\cite{roschewitz_automatic_2023}) effectively mitigate effects of acquisition shifts on model performance but will fail in the case of prevalence shift. The difficulty lies in the fact that a change in image characteristics may cause similar changes in the distribution over model outputs as a change in disease prevalence \cite{roschewitz_automatic_2023}, and determining the cause of an observed shift can be challenging. Despite its importance, automatic dataset shift identification has remained an open problem.

In this work, we address this issue by proposing a dataset shift \textit{identification} framework capable of identifying the root cause of the underlying shift, effectively separating (i) prevalence shift, (ii) covariate shift and (iii) mixed shift (both prevalence and covariate shifts). To the best of our knowledge, this is the first framework able to identify the type of test-time shifts in an unsupervised manner for imaging data, beyond solely detecting shifts. An in-depth evaluation across three different clinical applications (chest radiography, digital mammography, and retinal fundus images) on five types of real-world dataset shifts demonstrates that our framework accurately distinguishes between prevalence shifts, covariate shifts, and mixed shifts across various scenarios.

\section{Background}
\subsection{Definitions of prevalence and covariate shift}
\label{sec:shift_definition}
\index{prevalence shift}
\index{covariate shift}
Formally, let $X$ denote the input image and $Y$ denote the target (e.g. disease label). \emph{Label shift} (or prevalence shift) occurs when label distribution changes across domains, i.e. $P_{ref}(Y) \ne P_{test}(Y)$, while the conditional distributions, i.e. $P_{ref}(Y|X) = P_{test}(Y|X)$ are preserved, where $P_{ref}$ and $P_{test}$ denote distributions on reference and target domains respectively. Conversely, \emph{covariate shift} occurs when $P_{ref}(X) \ne P_{test}(X)$, while conditional distributions are preserved~\cite{murphy_probabilistic_2023}. Acquisition and subpopulation shifts are cases of covariate shifts as they directly affect image appearance.

\subsection{Dataset shift detection methods}
\label{sec:shift_details}
\index{Black Box Shift Detection (BBSD)}
\index{Dataset shift detection}
Several paradigms have been proposed for dataset shift detection. The simplest method to implement consists of comparing distributions of a classifier's outputs between the reference and test domain, proposed by Rabanser et al. \cite{rabanser_failing_2019}, and referred to as \emph{Black Box Shift Detection} (BBSD). In detail, softmax model outputs are collected for all samples in the reference and test sets. Then, for each class, a separate univariate Kolmogorov-Smirnov (K-S) test is run to determine if the class-wise predicted probabilities distributions differ between reference and test domain, the overall significance of the shift is then determined after applying Bonferroni correction~\cite{dunn_multiple_1961} for multiple testing. 

In the same study, Rabanser et al. \cite{rabanser_failing_2019} also proposed another type of shift detector where reference and test data input distribution are compared using a feature-based approach. In this test, the input sample (image) first gets projected to a smaller dimension, e.g. through a pretrained neural network encoder and the shift is then measured using the \emph{Maximum Mean Discrepancy}\index{Maximum Mean Discrepancy} permutation test originally proposed by Gretton et al. \cite{gretton_kernel_2012}. The Maximum Mean Discrepancy measures the distance between two distributions $P$ and $Q$ based on the distance between their mean embeddings. An unbiased estimate of the square of the MMD statistic can be computed via:
\begin{equation}
     \widehat{MMD}^2=\frac{1}{m^2-m}\sum_{i=1}^{m}\sum_{j \ne i}^{m}\kappa(z_i, z_j) + \frac{1}{n^2-n}\sum_{i=1}^{n}\sum_{j \ne i}^{n}\kappa(z'_i, z'_j) - \frac{2}{mn}\sum_{i=1}^{m}\sum_{j=1}^{n}\kappa(z_i, z'_j),
\end{equation} where $\{z_i\}_{i=1}^{m}\sim P$, and $\{z'_i\}_{i=1}^{n}\sim Q$ and $\kappa$ is a kernel over the embedding space. The p-value can then be obtained using a permutation test. In Rabanser et al. \cite{rabanser_failing_2019}, they proposed to use the RBF kernel $\kappa(z,\tilde{z})=e^{-\frac{1}{2\sigma}||z-\tilde{z}||^2}$, setting $\sigma$ as the median distance between all samples. An alternative way would be to explicitly learn the kernel, also known as `deep kernels'~\cite{liu_learning_2020}. The disadvantage of this approach is that the kernel needs to be learned for every single test set for which one wishes to run shift detection. In this work, we use the RBF kernel. 

Another existing shift detection approach consists of training a domain classifier to classify samples between the reference and test domains~\cite{cheng_classification_2022,lopez-paz_revisiting_2017,jang_sequential_2022} and using the accuracy of this classifier as a proxy for measuring distances between distributions. One drawback of this approach is its high computational cost: for every test set, a new domain classifier must be trained, which is highly impractical in continuous monitoring scenarios. Hence, we here focus on output-based (BBSD) and feature-based (MMD) shift detection methods, as these do not require training of additional models at test-time.

In the medical imaging domain, all these shift detectors were benchmarked by Koch et al.~\cite{koch_distribution_2024} for their ability to detect a large variety of real-world shifts in the context of diabetic retinopathy grading models. Both the domain classifier and the BBSD approach were shown to successfully detect the tested shifts, their sensitivity depending on the number of test samples, whereas MMD-based tests were found to be less accurate at detecting shifts compared to BBSDs and domain classifiers. Given these results, and for computational efficiency, we herein focus on shift detection methods that do not require the training of additional models at test-time.

\section{Methods and experimental setup}
\label{sec:shift:method}
\subsection{Dataset shift identification pipeline}
\label{sec:shift:shift_identification}
\index{dataset shift identification}
\begin{figure}[h]
    \centering
    \includegraphics[width=0.95\textwidth]{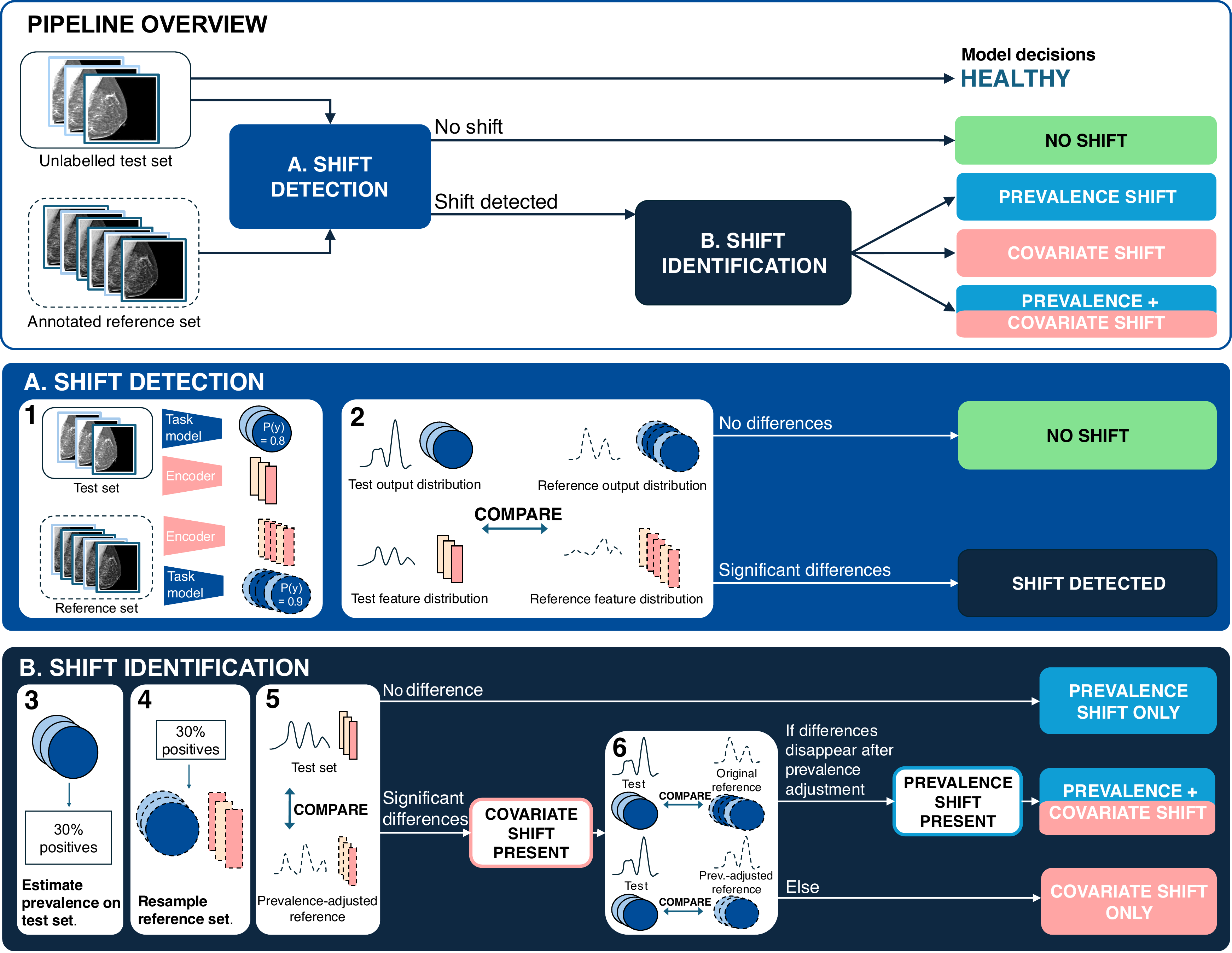}\\
    \caption[Overview of the proposed dataset shift identification pipeline]{\textbf{Overview of the proposed dataset shift identification pipeline}. We leverage both task model outputs and features from self-supervised encoders for detecting and identifying dataset shifts. Contrarily to previous works, we do not simply detect the \emph{presence} of shifts but also add a second step able to \emph{identify} the nature of the shift. Our method effectively separates cases of (i) prevalence shift (a change in label distribution), (ii) covariate shift (a change in image characteristics) and (iii) covariate and prevalence shift (both).}
    \label{fig:shift:identification}
\end{figure} 

Here, we propose a framework for identifying whether dataset shift is caused by prevalence shift, by covariate shift or by a mix of both. The approach is divided into two stages, as depicted in \cref{fig:shift:identification}:

\begin{enumerate}
    \item \textbf{Standard dataset shift detection} to separate the `shift' from the `no shift' cases (\cref{fig:shift:identification}, A). For this step, we use a dual detection approach, combining signals from task model outputs and features from self-supervised (SSL) encoders to detect shifts. In detail, we first independently run the BBSD and the MMD shift detection tests; this yields $C$ p-values for the BBSD test (one per class) and one p-value for the MMD permutation test. We then apply Bonferroni correction on the $C+1$ p-values to get the overall significance.\index{Black Box Shift Detection (BBSD)} 
    \item \index{Prevalence shift adaptation}If a shift is detected in the first step, we then proceed to \textbf{shift identification}. This process starts with estimating the prevalence in the test set (\cref{fig:shift:identification}, B.3). For this, we can leverage the prevalence shift adaptation literature, where various methods have been proposed to estimate the density ratio:
    \begin{equation}
        w := \frac{P_{ref}(Y)}{P_{test}(Y)}
    \end{equation}
    with $w\in\mathbb{R}^C$ \cite{alexandari_maximum_2020,saerens_adjusting_2002,wen_class_2024}. Here, we use the state-of-the-art `class probability matching with calibrated network' (CPMCN) method by Wen et al. \cite{wen_class_2024} to estimate this ratio and the test set prevalence. In CPMCN, the density ratio is estimated by: 
    \begin{equation}
        \hat{w} := \arg\min_{w\in\mathbb{R}^C}\sum_{i=1}^{C}\left|\hat{P}_{ref}(Y=i) - \frac{1}{m}\sum_{x \in D_{t}}\frac{\hat{p}(i|x)}{\sum_j^C w_j \hat{p}(j|x)}\right|^2
    \end{equation}
    with $m$ the number of samples in the test set $D_{t}$, $\hat{P}_{ref}(Y=i)$ is the empirical proportion of samples with class $i$ in the reference set, and $\hat{p}(i|x)$ is the probability predicted by the model for class $i$ given sample $x$. Given this estimated ratio $\hat{w}$, we can easily recover the estimated label distribution on the test domain 
    \begin{equation}
    \hat{P}_{test}(Y=i) = \hat{w}_i \cdot \hat{P}_{ref}(Y=i), \forall i \in C
    \end{equation}
    We follow this method to estimate the label distribution in the test set in our shift identification module. Next, we resample the reference set to match this estimated prevalence (\cref{fig:shift:identification}, B.4). We then first compare feature distributions between prevalence-adjusted reference and test set (\cref{fig:shift:identification}, B.5): if differences are no longer significant after adjusting the prevalence, the shift is attributed to prevalence shift. Conversely, if differences persist after adjusting the prevalence, then covariate shift is necessarily present. In this case, we compare model output distributions with BBSD to determine whether prevalence shift is also present (\cref{fig:shift:identification}, B.6). Precisely, if there were significant differences in model output distributions before adjusting the prevalence, but this shift disappears after adjusting the prevalence, we know that prevalence shift is also responsible for the observed shift, in this case we conclude that the observed shift is a case of mixed shift  (prevalence + covariate shift). Else, we conclude that the shift is attributed to covariate shift only.
\end{enumerate}

\subsection{Datasets}
\label{sec:shifts:datasets}
We evaluate our methods on four different datasets covering three different imaging modalities: (i) chest radiography, (ii) mammography, and (iii) fundus images. 

For chest radiography, we use two public datasets for our analysis: the \textbf{RSNA Pneumonia} dataset~\cite{shih_augmenting_2019}, a subset of the NIH Chest-Xray8 dataset~\cite{wang_chestx-ray8_2017} manually relabelled by expert radiologists for presence of pneumonia-like opacities. We use the original metadata from the NIH Chest-Xray8 dataset~\cite{wang_chestx-ray8_2017} to retrieve patient gender. We also use \textbf{PadChest}~\cite{bustos_padchest_2020} a larger dataset containing multiple disease labels extracted from radiology reports. We here focus on the pneumonia label. Importantly, this dataset contains important metadata such as scanner information or the gender of the patient, allowing us to generate a wide range of shifts. For mammography, we use the \textbf{EMBED} dataset~\cite{jeong_emory_2023} a large mammogram dataset collected in the US, on six different scanners. Finally, for fundus imaging, we create \textbf{RETINA} a multi-domain dataset by combining three different public datasets: the Kaggle Diabetic Retinopathy Detection~\cite{dugas_diabetic_2015} dataset, the Kaggle Aptos Blindness Detection dataset~\cite{karthik_aptos_2019} and the Messidor-v2 dataset~\cite{decenciere_feedback_2014}. These datasets cover different regions of the world (India, France, US) but also with varying image acquisition devices: images from the Messidor-v2 are high-quality images, while many images in the Kaggle datasets are of lower quality (including phone pictures). Creating this multi-centre dataset allows us to simulate various domain shifts by varying the proportion of data from each `site' (original dataset source). The task of interest for fundus images here is binary diabetic retinopathy (DR) classification for fundus images, where we classify images between referable DR (grades 2,3,4) and healthy/non-referable DR (grades 0,1).

\subsection{Shift generation details}
\label{sec:shift_generation}

For every dataset, we study different types of shifts at different levels of intensity. To study prevalence shift detection, we associate each dataset with a downstream task. For chest radiography datasets we focus on pneumonia detection, for mammography on breast density assessment (4 classes), and for retinal images on binary diabetic retinopathy classification. We simulate various levels of prevalence shift by resampling the test set according to specific label distributions. Then, we study various types of covariate shifts. For PadChest, we study gender shifts by varying the proportion of female patients in the test set. Moreover, PadChest contains scans acquired with two scanners, `Phillips' (40\%) and `Imaging' (60\%). This allows the simulation of different levels of acquisition shift by varying the proportion of Phillips scans in the test set. Similarly, for EMBED~\cite{jeong_emory_2023}, we study acquisition shift\index{Acquisition shift} by varying the distribution of scanners in the test set. This dataset offers a complementary view to PadChest, with a multi-class task of interest and providing even more flexibility for simulating diverse acquisition shifts (six scanners). Note that, in EMBED, each exam comprises four mammograms (left/right breasts and MLO/CC views). We excluded all exams that did not contain exactly four images, kept exactly one exam per patient, and ensured that test set sampling was done at the exam level. Finally, for the RETINA dataset, we simulate covariate shifts by varying the proportion of samples coming from each underlying dataset (Aptos, Kaggle DR and Messidor).

\subsection{Implementation details}
\label{sec:evaluation_details}

To evaluate the shift identification accuracy of the proposed framework, we follow standard evaluation practices from the dataset shift detection literature. Specifically, we repeat the following process 200 times: (i) sample a subset of size $N_{test}$ from the test split according to the shift of interest and sample the reference set from the validation split, (ii) run the shift identification test, and (iii) record whether the shift is correctly identified. The final shift identification accuracy is computed as the proportion of times the shift is correctly identified out of all the bootstrap samples. For both chest radiography datasets, we used  $N_{ref}=2,000$ images and $N_{test} \in \{100,250,500,1000\}$ images. For the RETINA dataset, we used  $N_{ref}=1000$ images and $N_{test} \in \{100,250,1000\}$ images. The case of EMBED is slightly different as we need to ensure that images belonging to the same exam are always sampled simultaneously. Hence, for this dataset sampling of all the reference sets, shifted test sets (and permutations in the MMD test\footnote{Keeping the underlying structure of the data during permutation tests is important to avoid inflated type I error~\cite{churchill_naive_2008}.}) is done at the exam level and not at the image level. We assumed $N_{exams,ref}=1000$ exams (i.e $4000$ images) and $N_{exams,test} \in \{50,100,250\}$.  
 
All task models and encoders used in this study are ResNet-50~\cite{he_deep_2016}. For each dataset, we use the task model for the BBSD test. For the MMD detection test, we first extract the embeddings from the last layer of the encoder, then project them onto the 32-first principle components and then use the RBF kernel. We compare various encoders for feature extraction, in particular, encoders trained in a self-supervised manner. Self-supervised (SSL) encoders were trained using the SimCLR~\cite{chen_simple_2020} objective. Additionally, for the RETINA dataset, we compare with RetFound~\cite{zhou_foundation_2023} a publicly available self-supervised foundation model trained with the Masked Auto Encoder \cite{he_masked_2022} objective on a large set of retinal images, based on a vision transformer architecture \cite{dosovitskiy_image_2020}.

\section{Results}

\subsection{Different shifts require different detectors} 

Prior to diving into shift \emph{identification}, we first investigate which types of shifts are successfully \emph{detected} by prominent dataset shift detection methods. We compare two families of shift detectors: model output-based (BBSD) and feature-based detectors (MMD). We additionally test a dual approach that combines both approaches for improved shift detection (`Duo', as described in \cref{sec:shift:shift_identification}).\index{Black Box Shift Detection (BBSD)} 

\begin{figure}
    \centering
    \includegraphics[width=0.97\textwidth]{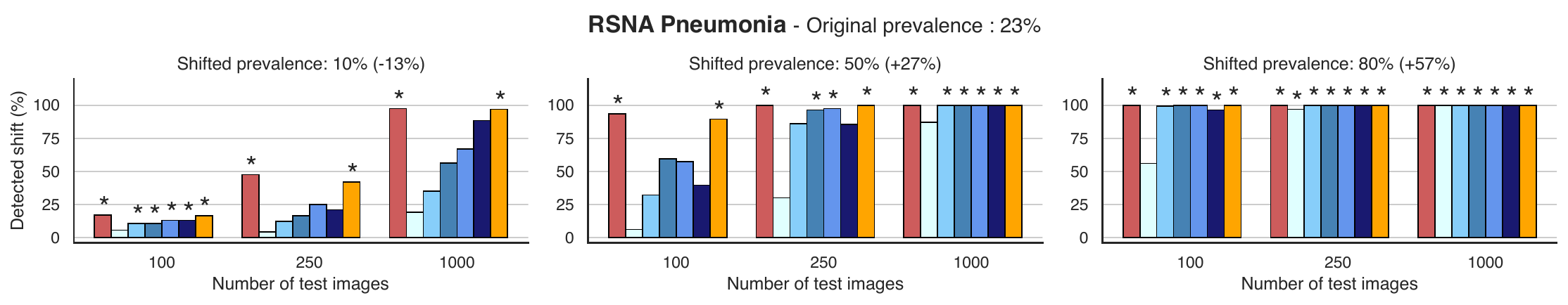}
    \includegraphics[width=0.97\textwidth]{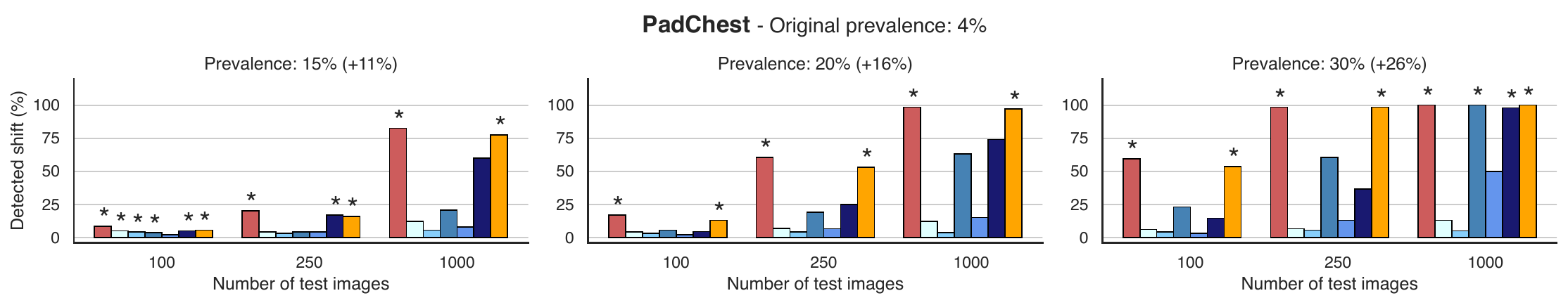}
    \includegraphics[width=0.97\textwidth]{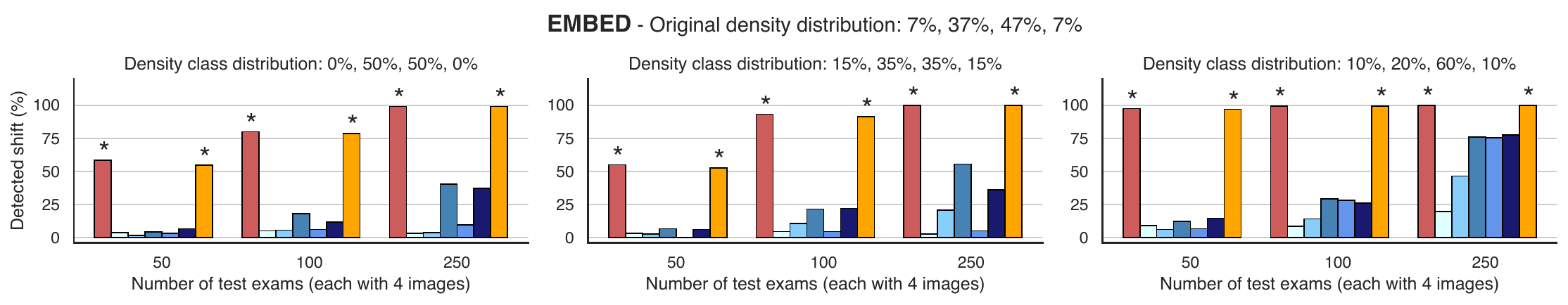}
    \includegraphics[width=0.97\textwidth]{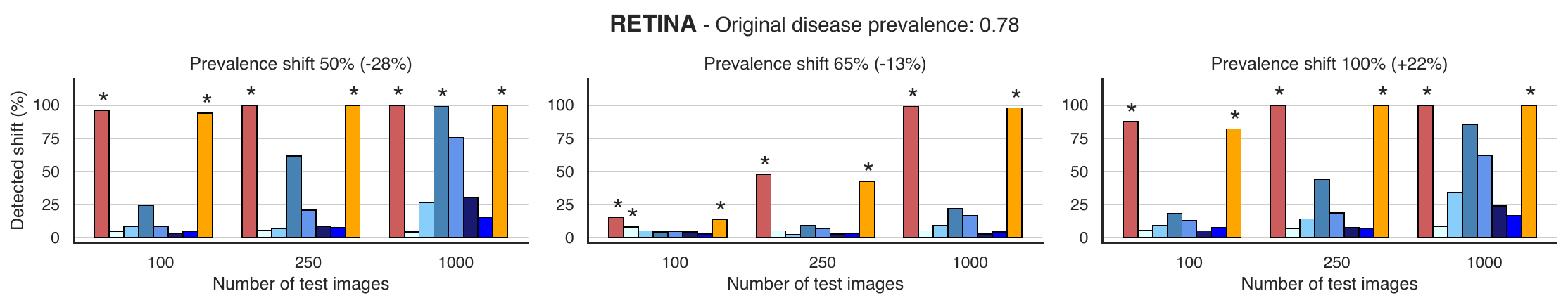}
    \includegraphics[width=0.97\textwidth]{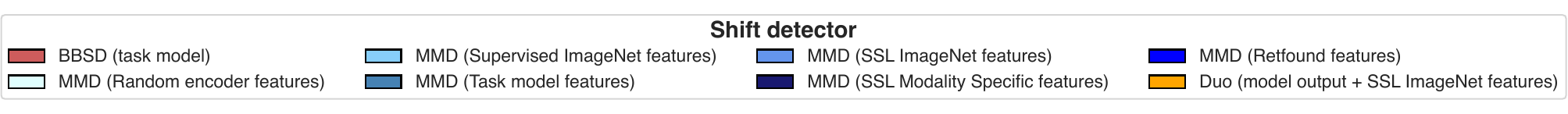}
    \caption[Prevalence shift: shift detectors comparison]{\textbf{Prevalence shift: shift detectors comparison}. We report the shift detection rate over 200 bootstrap samples. Results show that output-based detection is best for detecting prevalence shifts. For each combination of test set size and type of shift, the detector with the highest detection rate, as well as all detectors not significantly different from the best, are denoted with an asterisk $*$ (where significance is measured by Fisher's exact test, at level .05, with Bonferonni correction applied for multiple testing).}
    \label{fig:prev_detection}
\end{figure}

\begin{figure}
    \centering
    \includegraphics[width=0.97\textwidth]{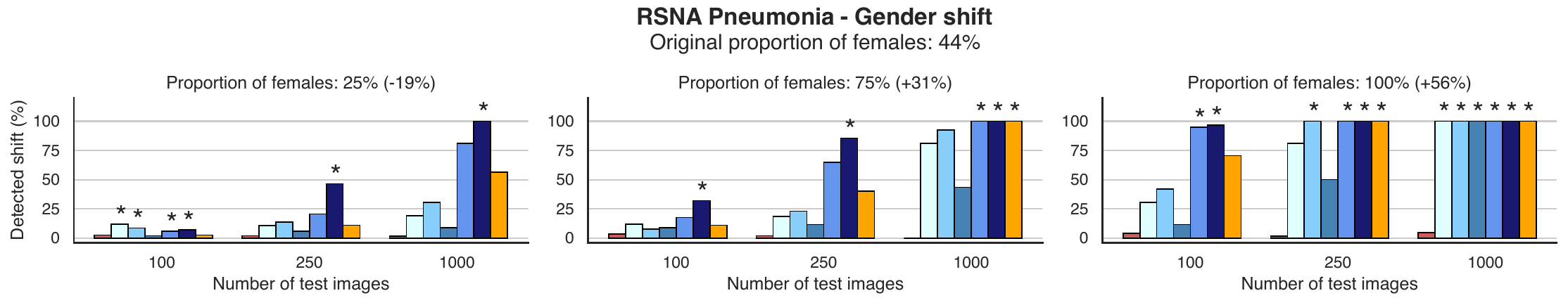}
    \includegraphics[width=0.97\textwidth]{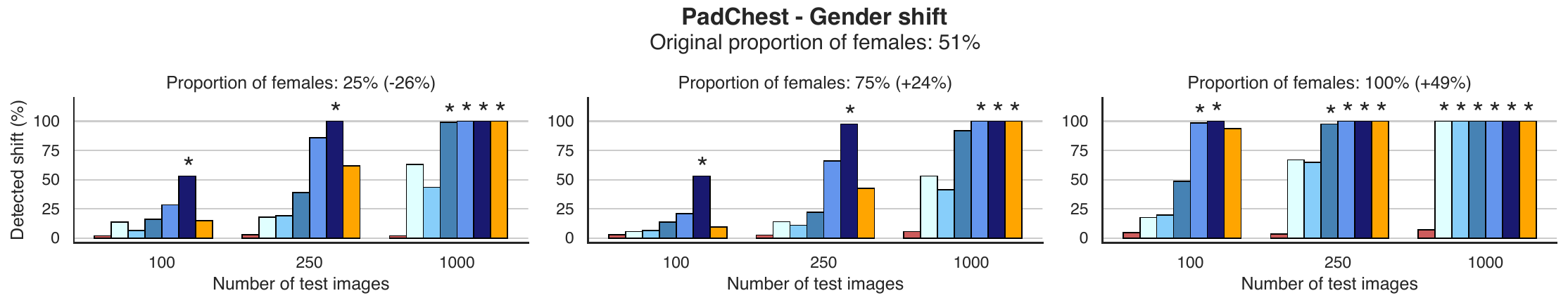}
    \includegraphics[width=0.97\textwidth]{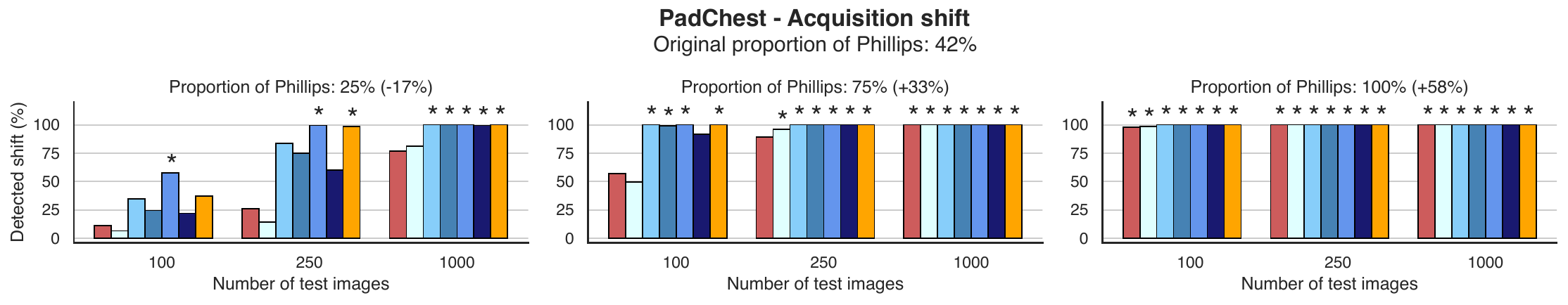}
    \includegraphics[width=0.97\textwidth]{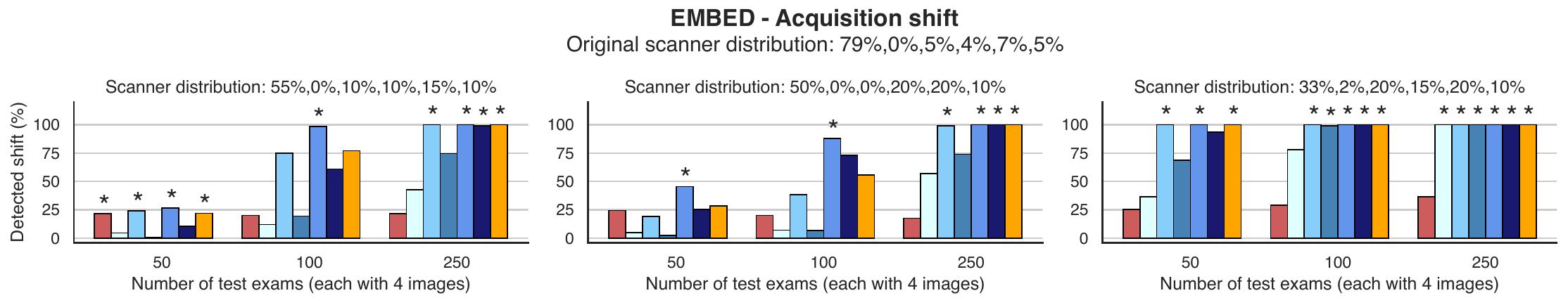}
    \includegraphics[width=0.97\textwidth]{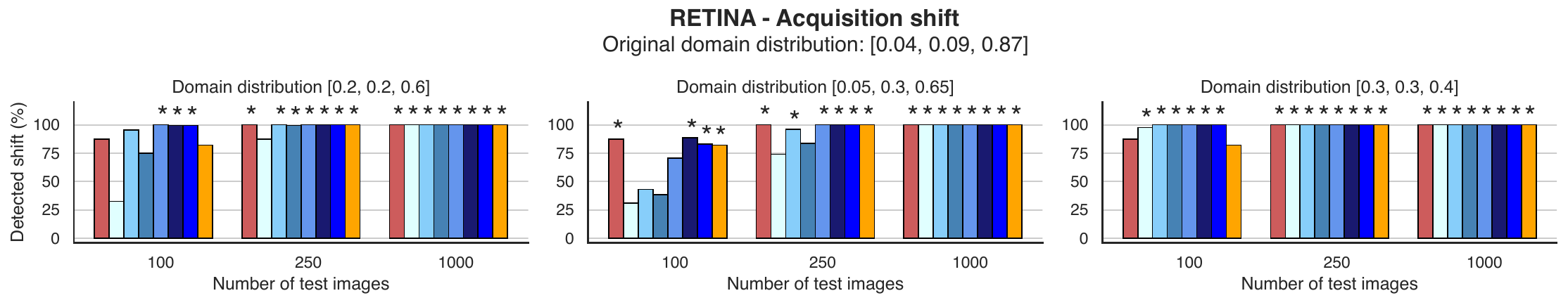}
    \includegraphics[width=0.97\textwidth]{figures/legend_detection.pdf}
    \caption[Covariate shifts: shift detectors comparison]{\textbf{Covariate shifts: shift detectors comparison}. We studied two sub-types of covariate shifts: subpopulation shift (top two rows) and acquisition shift\index{Acquisition shift} (bottom three rows). We report the shift detection rate over 200 bootstrap samples. Results show that feature-based detection is best for this type of shift. For each combination of test set size and type of shift, the detector with the highest detection rate, as well as all detectors not significantly different from the best, are denoted with an asterisk $*$ (where significance is measured by Fisher's exact test, at level .05, with Bonferonni correction applied for multiple testing).}
    \label{fig:covariate_detection}
\end{figure}

For feature-based shift detection, any pretrained network could be used as feature extractor. A perhaps obvious choice is to simply use the encoder from the task-specific classification model. However, this may not be the best choice as learned features will be heavily skewed towards encoding characteristics specifically relevant to that task as opposed to encoding more generic image representations sensitive to data distribution changes~\cite{zamzmi_out--distribution_2024}. Hence, we here explore the potential of SSL image encoders for shift detection. Indeed, encoders trained in a self-supervised manner, i.e. without any labels, learn fine-grained representations effectively summarising all the information encoded in a given image, resulting in ideal candidates for generic shift detection. We compare the performance of feature-based shift detection for five different encoders (using the features obtained from the last layer of the neural network encoder). We compare: (i) \textbf{Random} a ResNet-50 encoder with random weights; (ii)  \textbf{Supervised ImageNet} where features are extracted using a ResNet encoder trained to perform classification on natural images from the ImageNet dataset; (iii)  \textbf{Task model}, the ResNet encoder from the task model used to perform the downstream classification task; (iv) \textbf{SSL ImageNet}  a self-supervised encoder trained on ImageNet data only; (v)  \textbf{SSL Modality Specific} a self-supervised encoder trained on the same modality as the test datasets. Additionally, for the RETINA dataset, we include a comparison using the  \textbf{RetFound} foundation model as a feature extractor~\cite{zhou_foundation_2023}.

\paragraph{Output- and feature-based detectors detect different shifts.} \cref{fig:prev_detection,fig:covariate_detection} show the shift detection rates for every dataset-shift combination.  Across all datasets, a clear pattern appears. For prevalence shifts (\cref{fig:prev_detection}), output-based shift detection performs significantly better than all feature-based tests. This is intuitive as the shift is directly related to the downstream prediction task. A change in prevalence should directly be reflected by a change in the distribution of task model outputs. For covariate shifts, results are very different, regardless of whether we look at acquisition or subpopulation shifts (see \cref{fig:covariate_detection}). For these shifts, most feature-based detectors perform substantially better than output-based detectors. Output-based shift detection fails to detect gender shifts on radiography datasets (less than 5\% of shifts detected) and performs substantially worse than feature-based tests on acquisition shifts across all levels of shifts and datasets. For example, on EMBED with 250 test exams, the output-based shift detector only detects 25\% of acquisition shifts, whereas feature-based detectors detect at least 80\% of shifts (except for the random encoder). 

\paragraph{Self-supervised encoders can detect subtle covariate shifts.}
\label{sec:encoders}

Our results show that the effectiveness of feature-based dataset shift detectors is highly dependent on the choice of the encoder used to extract the features (see differences between feature-based detection rates in shades of blue in  \cref{fig:covariate_detection}). The results indicate that for optimal detection of more subtle shifts (e.g. gender shifts), it is important to use an encoder trained in a self-supervised manner. The task model, for example, under-performs for gender shift detection on chest radiography datasets, and so does the encoder trained in a supervised manner on ImageNet data, particularly visible for milder shifts and smaller test set sizes. For example, both supervised encoders fail to detect the shift from 44\% to 25\% of females in the RSNA dataset, even with large test sizes (\cref{fig:covariate_detection} top row). Encoders trained in a self-supervised manner (SSL ImageNet and SSL Modality Specific) detect gender shifts with significantly higher sensitivity, exhibiting an average sensitivity of 85\% across all gender shifts with a test set size of 250 and 100\% for a test set size of 1000. Similarly, for acquisition shifts, results show that SSL encoders offer substantially better detection rates than their supervised counterparts. The SSL model trained on ImageNet data was particularly effective and, in some cases, even better than the modality-specific SSL models for PadChest and EMBED, especially for subtle shifts and in the low test data regime. On the RETINA dataset, self-supervised encoders also all outperform their supervised counterparts, with no significant differences between self-supervised encoders.

\paragraph{Combining output- and feature-based detection.} Building on the previous findings, we evaluate a dual detection approach combining responses from output-based and feature-based shift detectors, using self-supervised features for more robust shift detection. Results in \cref{fig:prev_detection,fig:covariate_detection} demonstrate that the proposed `Duo' detector (in orange) performs best overall across shifts and datasets, regardless of the shift severity. For example, with the duo detector and test set size of 1000 samples, the detection accuracy is $>$80\% for nearly every shift, across all datasets. On the contrary, the other detectors respectively fail either on prevalence shift (with feature-based detectors detecting less than 50\% prevalence shift cases averaged over datasets) or on covariate shift (where output-based detection detects less than 5\% of gender shifts for radiography and less than 25\% of acquisition shifts for EMBED). We employ this dual approach for the detection module of our shift identification pipeline.

\begin{figure}
    \centering
    \includegraphics[width=\textwidth]{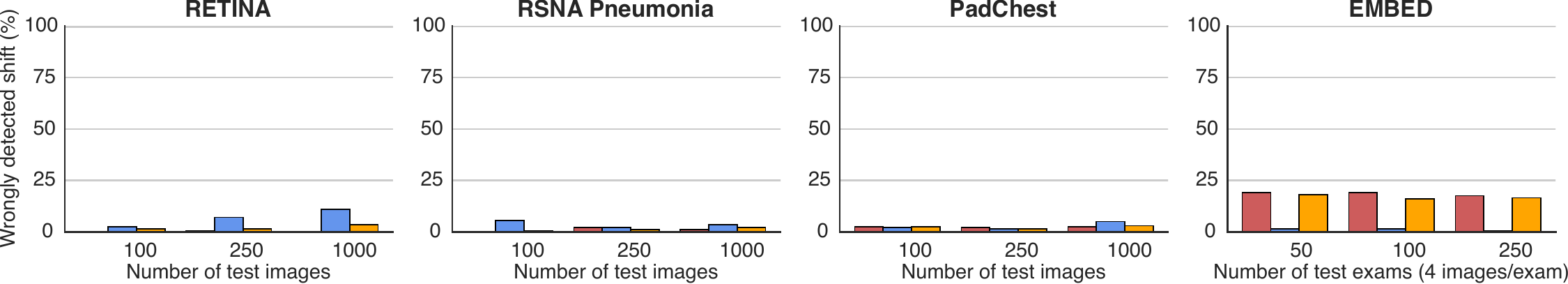}
    \includegraphics[width=.9\textwidth]{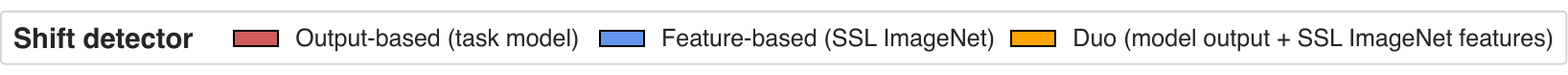}
    
    \caption[False detection rate: shift detectors comparison]{\textbf{False detection rate: shift detectors comparison}. Percentage of shift detected when resampling the test set without shift.}
    \label{fig:no_shift}
\end{figure}

For completeness, in \cref{fig:no_shift} we report the false shift detection rate of shift detectors when generating test sets without shifts. We verify that the false-positive rates of the output-based, feature-based and Duo detectors stay low across datasets and detectors, we find that false detection rates hover around the expected type-I error of the statistical tests 5\% for most datasets and detection methods.  The only exception is EMBED, where false positive rates are a little higher than expected for the output-based (and hence Duo) detector. This is due to the fact that - for this highly imbalanced classification problem - small randomly resampled test sets may lead to small differences in the observed cumulative distribution of predictions, even in the absence of a shift in the test set generation process.

\subsection{Shift identification performance} 
To perform shift identification, we leverage the fact that output-based and feature-based shift measures detect different types of shifts to precisely identify the shift present in the test dataset, following the decision logic detailed in \cref{sec:shift:shift_identification} and illustrated in~\cref{fig:shift:identification}.

\begin{figure}
    \centering

    \includegraphics[width=0.42\textwidth]{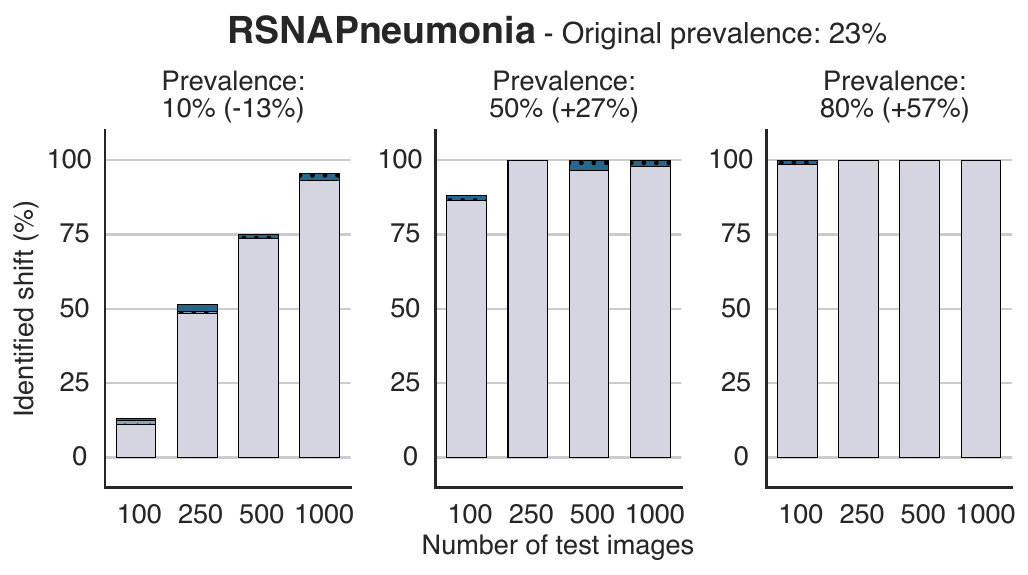}\hspace{0.04\textwidth}
    \includegraphics[width=0.42\textwidth]{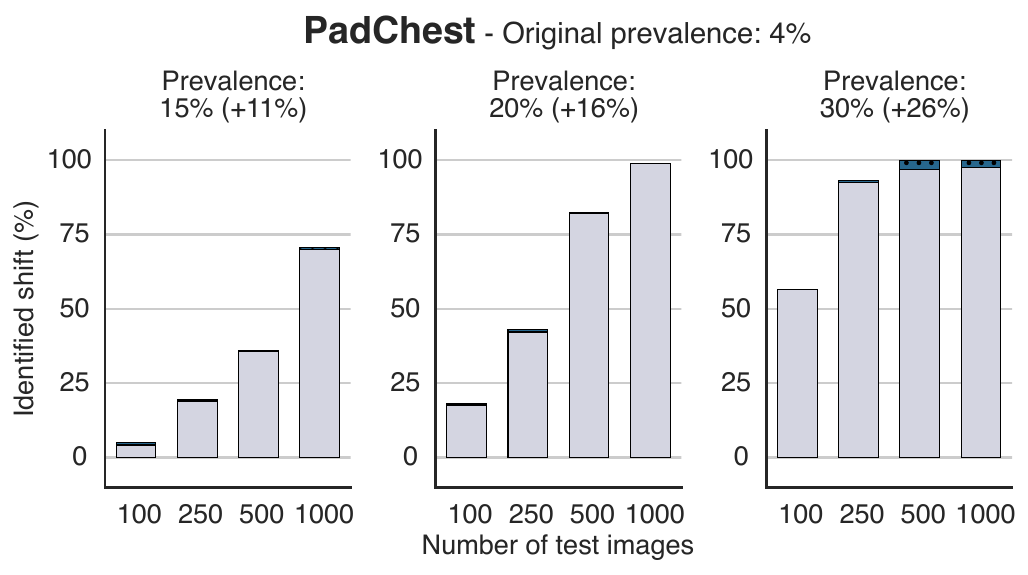}\\
    \vspace{3mm}
    \includegraphics[width=0.42\textwidth]{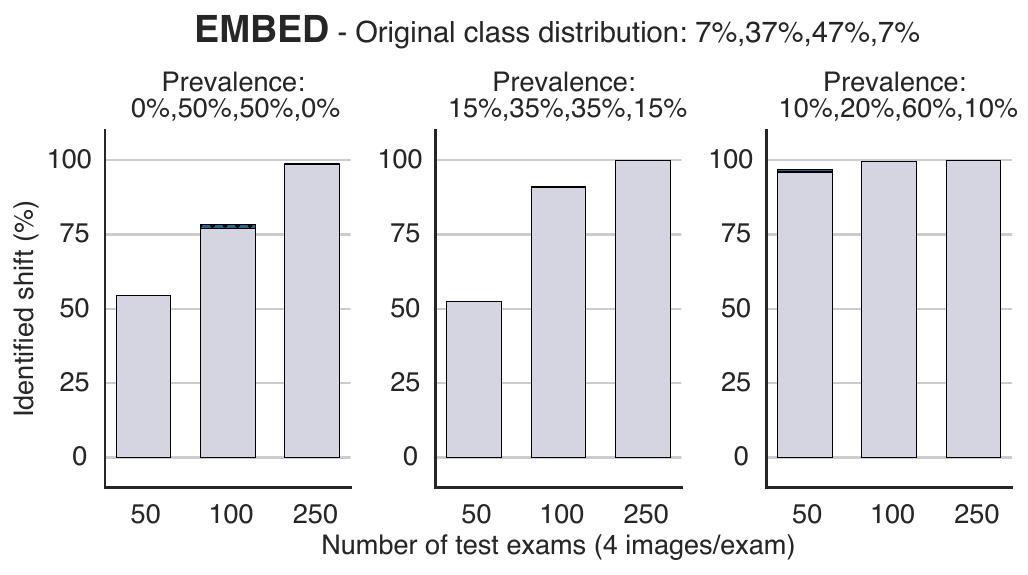}\hspace{0.04\textwidth}
    \includegraphics[width=0.42\textwidth]{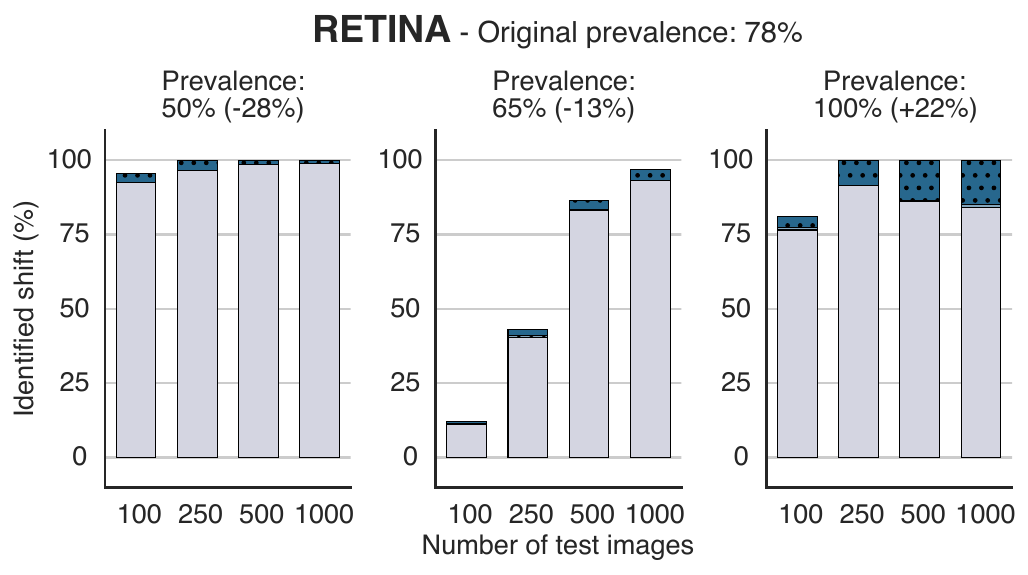}\\
    \vspace{3mm}
    \includegraphics[width=0.42\textwidth]{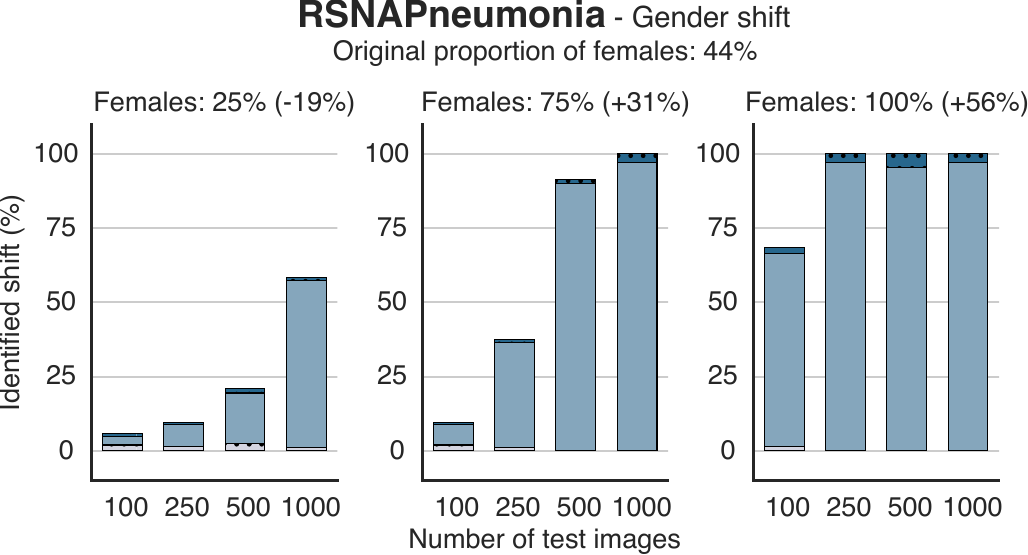}\hspace{0.04\textwidth}
    \includegraphics[width=0.42\linewidth]{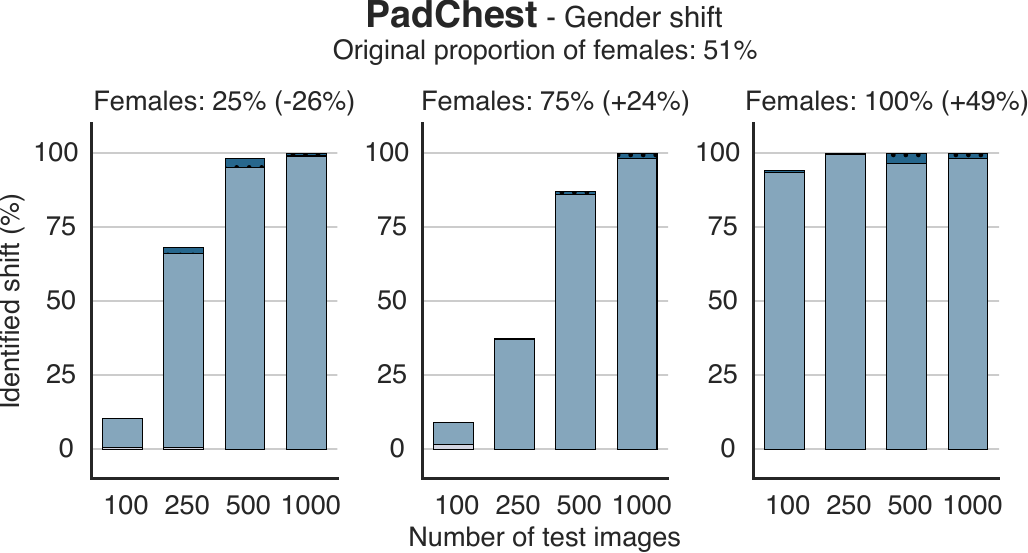}\\
    \vspace{3mm}
    \includegraphics[width=0.42\textwidth]{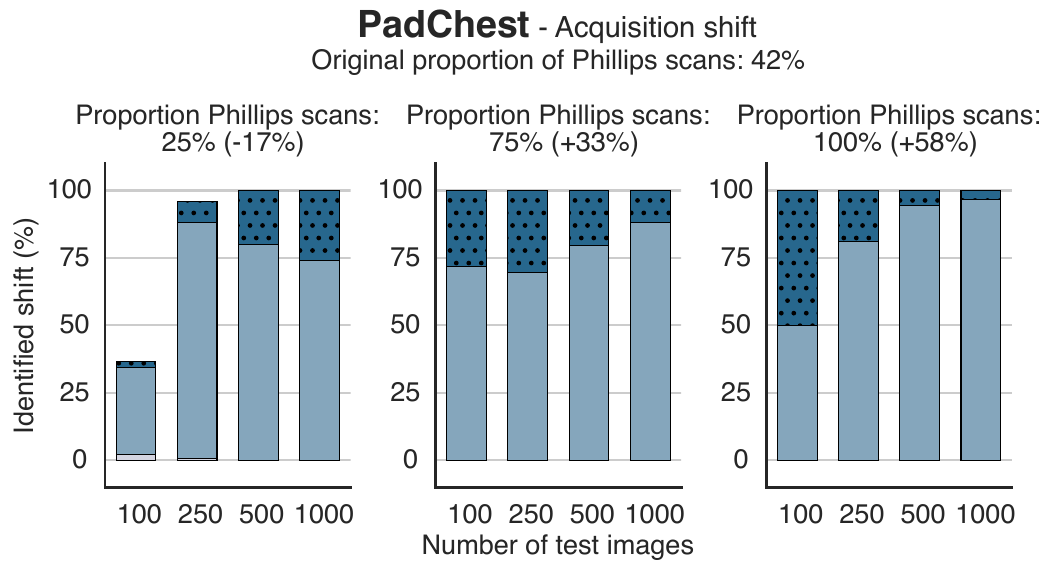}\hspace{0.04\textwidth}
    \includegraphics[width=0.42\textwidth]{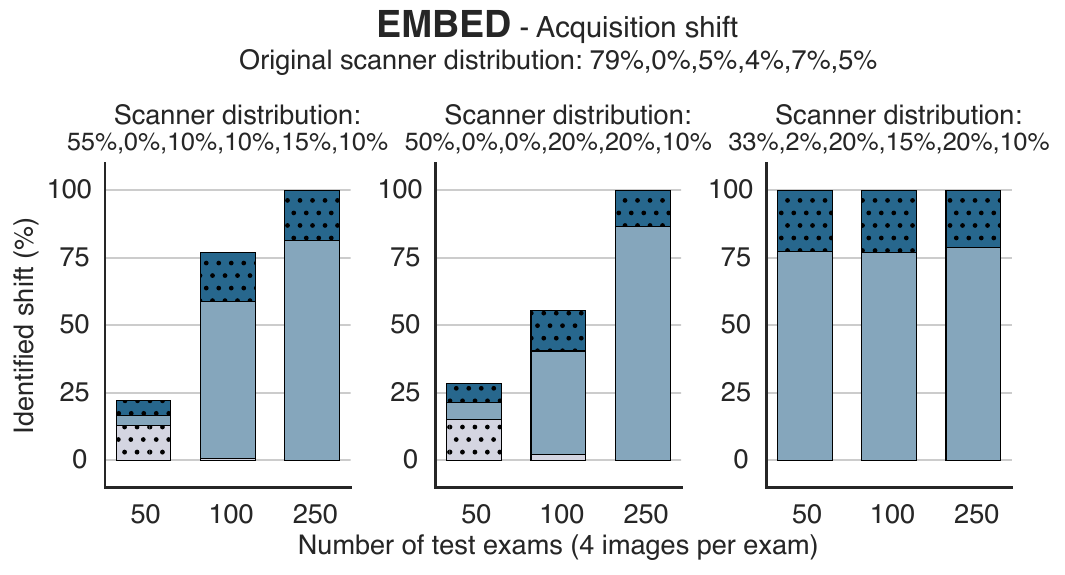}\\
    \vspace{3mm}
    \includegraphics[width=0.42\textwidth]{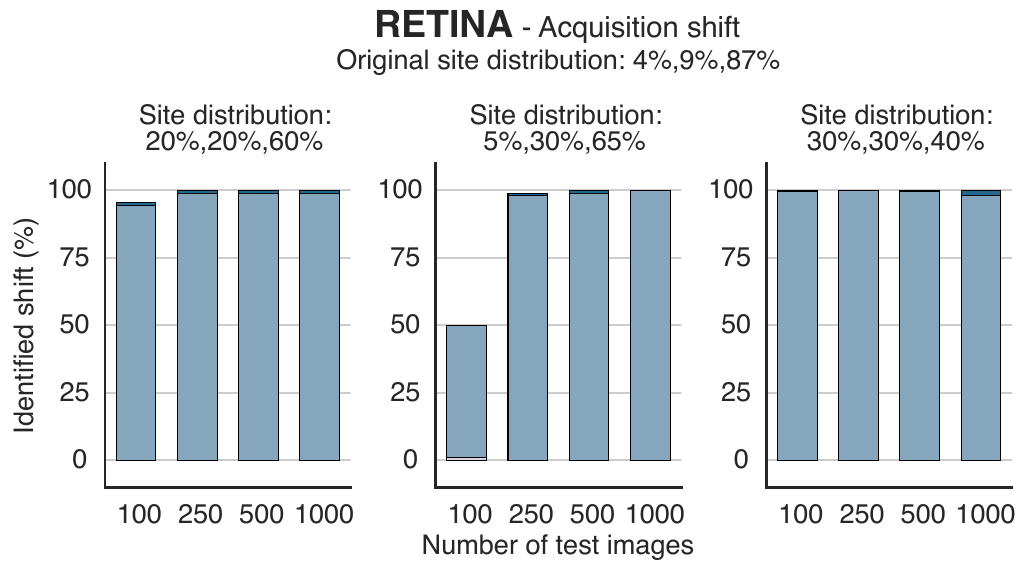}
    \includegraphics[width=\textwidth]{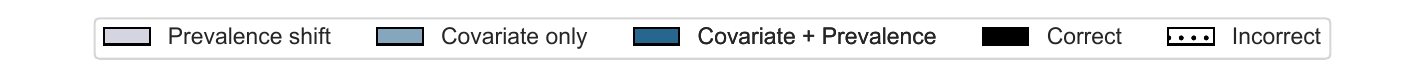}
    \caption[Shift identification accuracy: prevalence shifts and covariate shifts]{\textbf{Shift identification accuracy: prevalence shifts (top two rows) and covariate shifts (bottom three rows)}. Across all datasets, the shift identification framework is able to successfully detect and identify both prevalence shifts and covariate shifts with high accuracy. Identification accuracy is computed over 200 bootstrap samples.}
    \label{fig:identification_covariate}
\end{figure}

\begin{figure}[H]
    \centering
     \includegraphics[width=0.47\textwidth]{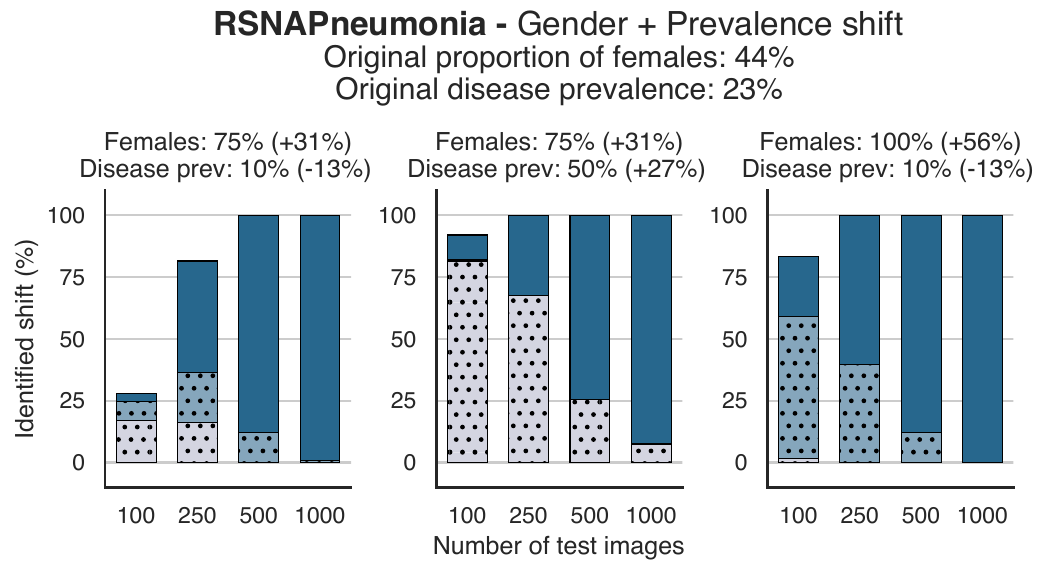}
     \hspace{0.03\textwidth}
\includegraphics[width=0.47\textwidth]{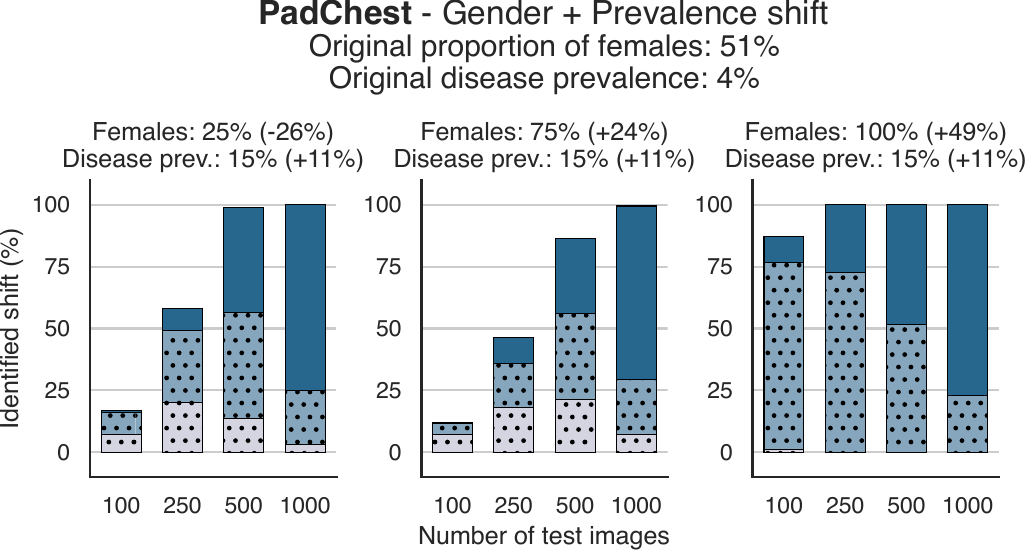}
    
\includegraphics[width=0.47\textwidth]{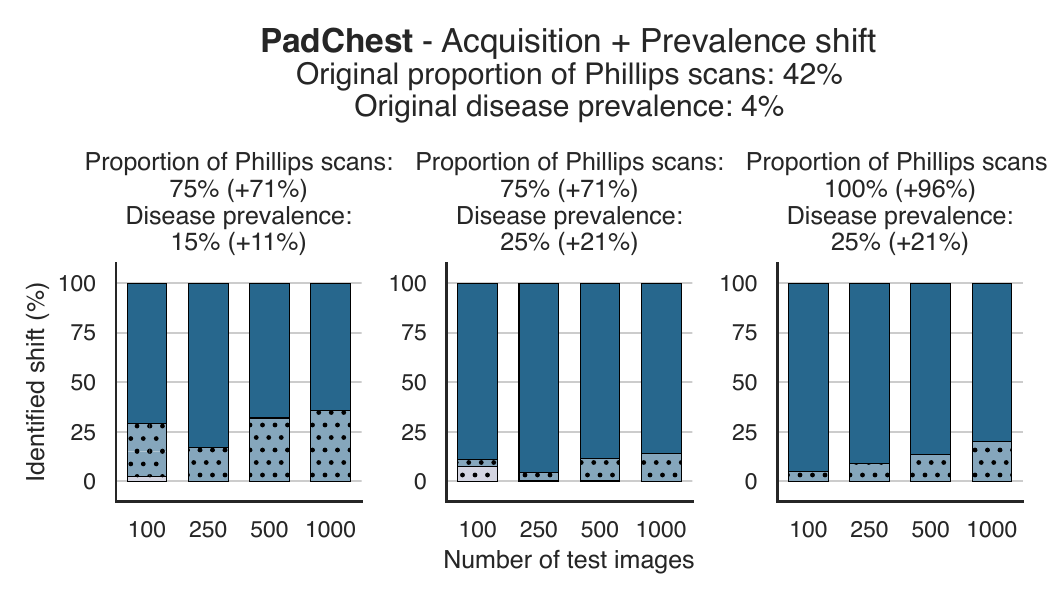}
     \hspace{0.03\textwidth}
\includegraphics[width=0.47\textwidth]{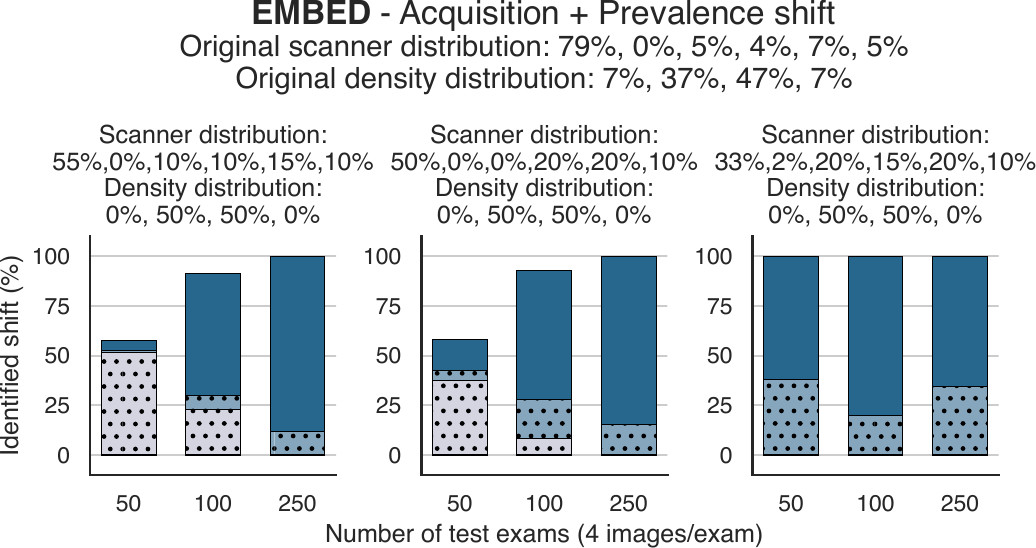}

\includegraphics[width=0.47\textwidth]{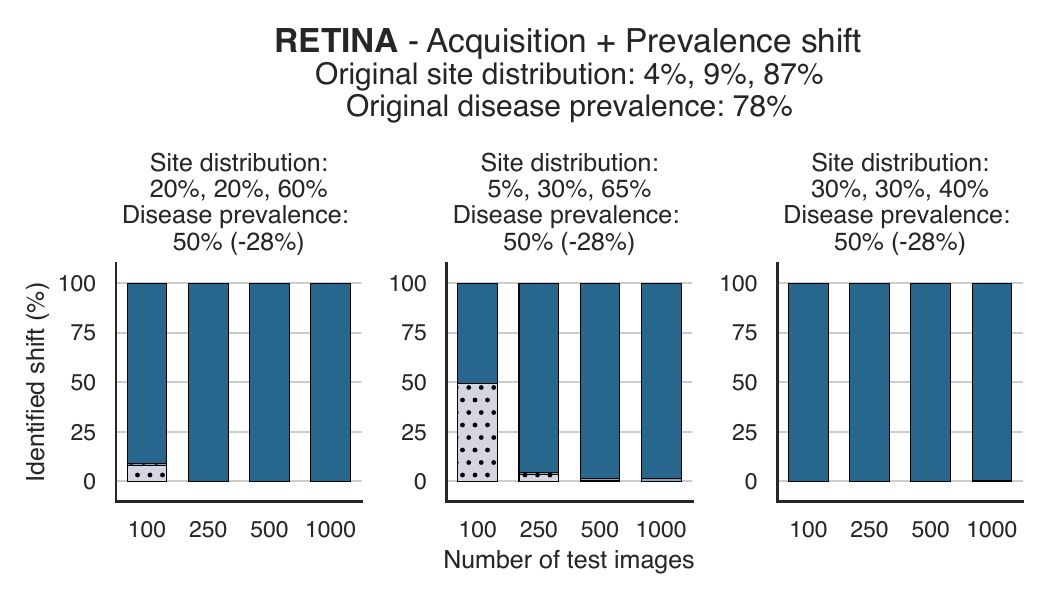} \\
     \includegraphics[width=\textwidth]{figures/legend_id.pdf}
    \caption[Shift identification accuracy: mixed covariate and prevalence shifts]{\textbf{Shift identification accuracy: mixed covariate and prevalence shifts}. (Top row) depicts mixed gender and prevalence shift, (bottom two rows) show mixed acquisition and prevalence shift. Across all datasets, the shift identification framework is able to successfully detect and identify presence of both shifts with high accuracy. Identification accuracy is computed over 200 bootstrap samples.}
    \label{fig:identification_mixed}
\end{figure}

In-depth evaluation results in \cref{fig:identification_covariate,fig:identification_mixed} demonstrate that our novel shift identification framework is capable of distinguishing between prevalence shifts, covariate shifts and mixed shifts with high accuracy across all datasets and types of shifts. Overall, more subtle shifts are best detected with larger test sets whereas larger shifts can be detected with smaller test sets. For prevalence shifts, the average shift identification rate, across datasets and shift levels, is 85\% with 500 test images and reaches 95\% with 1000 test images (\cref{fig:identification_covariate}, top two rows).

 Similarly, for covariate shifts caused by acquisition shifts, when using a test set size of 500 images (and 250 test exams on EMBED),  the identification accuracy is greater than 80\% for any shift level and dataset, with an average identification accuracy of 89\% across datasets and shift levels (\cref{fig:identification_covariate} bottom two rows). For the RETINA dataset, identification rates already reach 100\% with a test set as small as 250 images. When the covariate shift is induced by gender shifts, covariate shift is detected with identification accuracies greater than 90\% for both PadChest and RSNA Pneumonia, with a test set size of 500 images, for all but one shift level (\cref{fig:identification_covariate}). Results in~\cref{fig:identification_mixed}, show that the framework is also able to accurately distinguish between cases of covariate shift only and cases of covariate and prevalence shifts. For these mixed shifts,  shifts are identified with increasingly high accuracy as the test set size increases. With a test set of 1000 images, mixed gender and prevalence shifts are correctly identified as mixed shifts with an average accuracy of 97\% for RSNA Pneumonia and 75\% for PadChest. Mixed shifts induced by acquisition and prevalence shifts are detected with an average 100\% accuracy for the RETINA dataset, 77\% for PadChest and 79\% for EMBED, across shift levels with 1000 test images. The overall identification accuracy for mixed shifts across all shifts and datasets is 85\% (with 1000 test images).

\section{Discussion}
By analysing common dataset shift detection paradigms, we find that different types of shifts require different types of shift detectors. Our analysis also demonstrates the importance of the choice of encoders for feature-based dataset shift detection. In particular, we find that encoders trained in a self-supervised manner yield features with substantially higher shift detection power than supervised counterparts. Maybe surprisingly, our results show that generic encoders trained in a self-supervised manner on natural images (ImageNet) provide highly discriminative features for medical image dataset shift detection, transportable across datasets. Following these findings, we evaluate a new dual dataset shift detector, combining shift detection signals from task model outputs tests and features from self-supervised encoders. This approach outperforms existing shift detectors. The consistency of shift detection performance across various types of shifts is crucial as the nature of the shift is, by definition, unknown at test time. 

We take this combined approach a step further to perform fine-grained shift identification, showing high accuracy in correctly identifying the type of shift present in the test set across all modalities, tasks and various levels of shift intensity. Our results demonstrate the practical value of the shift identification method, which does not require any training at test time, nor any ground truth labels or annotations on the test domain data. The use of a readily available self-supervised encoder trained on ImageNet data for feature extraction dispenses us from training any additional model for shift detection and identification purposes. Combining signals from both feature-based and model output-based shift detectors yields reliable and consistent detection and identification across all types of shifts. Importantly, our method not only separates covariate shifts from prevalence shifts but also reliably detects when both types of shifts are present in the test set, rendering the proposed method applicable to many real-world deployment scenarios.

In practice, the proposed framework can be used as a continuous monitoring tool for any image-based, clinical AI model. Similar to other continuous monitoring tools \cite{merkow_chexstray_2023}, the shift identification framework would run on the stream of incoming data, collecting test data on a rolling time window. The reference data should be a small dataset representative of the expected data distribution on which the model has been validated. Our results show that with at most 2000 reference images, a test set of only 500 images suffices to yield high shift detection and identification accuracy across all shifts and datasets ($>80\%$). 

In terms of limitations, we note that in the case of covariate shift, on its own, the proposed shift identification framework does not allow for a more fine-grained identification of the origin of shift, e.g. the distinction between population and acquisition shifts\index{Acquisition shift}. To allow for an even more precise sub-type shift identification, integrating metadata statistics in the pipeline (e.g. as in multi-modal shift detection pipelines ~\cite{merkow_chexstray_2023}) could complement the framework. Nevertheless, it is important to highlight that relying solely on metadata monitoring only enables the detection of shifts affecting the specific attributes collected at test time. For example, statistics on patient population, such as age distribution, gender distribution, may be recorded at deployment time and used to detect \emph{some} sub-types of shifts. Should the metadata not be available at test time, one could use auxiliary models to predict attributes of interest from images directly (e.g. ethnicity \cite{gichoya_ai_2022}). However, solely relying on collected (or predicted) metadata may not capture all sources of covariate shifts and will not allow detection of prevalence shift. The proposed framework can help uncover shifts that are not detectable by means of simply tracking population metadata. Shift detection and identification can prompt further investigation and inform root cause analysis of AI performance degradation.

\subsection*{Acknowledgments}
M.R. is funded by an Imperial College London President’s PhD
Scholarship and a Google PhD Fellowship. R.M. is funded through the European Union’s Horizon Europe research and innovation programme under grant agreement 10108030. C.J. is supported by Microsoft Research and EPSRC through the Microsoft PhD
Scholarship Programme. B.G. acknowledges support from the Royal Academy of Engineering as part of his Kheiron Medical Technologies/RAEng Research Chair in Safe Deployment of Medical Imaging AI.

\subsection*{Disclosure of interests}
B.G. is part-time employee of DeepHealth. No other competing interests.

\bibliographystyle{abbrv}
\bibliography{references}

\end{document}